# Process parameter optimization of Friction Stir Welding on 6061AA using Supervised Machine Learning Regression-based Algorithms

Eyob Messele Sefene[1*], Assefa Asmare Tsegaw[1], Akshansh Mishra[2]

[1] Bahir Dar Institute of Technology, Faculty of Mechanical and Industrial Engineering, Bahir Dar, 6000, Ethiopia

[2] Centre for Artificial Intelligent Manufacturing Systems, Stir Research Technologies, India

**Abstract:** The highest strength-to-weight ratio criterion has fascinated curiosity increasingly in virtually all areas where heft reduction is indispensable. Lightweight materials and their joining processes are also a recent point of research demands in the manufacturing industries. Friction Stir Welding (FSW) is one of the recent advancements for joining materials without adding any third material (filler rod) and joining below the melting point of the parent material. The process is widely used for joining similar and dissimilar metals, especially lightweight non-ferrous materials like aluminum, copper, and magnesium alloys. This paper presents verdicts of optimum process parameters on attaining enhanced mechanical properties of the weld joint. The experiment was conducted on a 5 mm 6061 aluminum alloy sheet. Process parameters; tool material, rotational speed, traverse speed, and axial forces were utilized. Mechanical properties of the weld joint are examined employing a tensile test, and the maximum joint strength efficiency was reached 94.2%. Supervised Machine Learning based Regression algorithms such as Decision Trees, Random Forest, and Gradient Boosting Algorithm were used. The results showed that the Random Forest algorithm yielded highest coefficient of determination value of 0.926 which means it gives a best fit in comparison to other algorithms.

## 1. Introduction

Before the invention of Friction Stir Welding, a mechanical fastener was the key element for joining aluminum and its alloy material in different aircraft structural parts. For instance, the Eclipse Aviation industry, to manufacture Eclipse 500 business class aircraft, approximately used 7,300 rivets per airframe (fuselage). After all this economic loss, Wayne Thomas and his friends at The Welding Institute (TWI) of Cambridge, UK, invented Friction Stir Welding in 1991 [1, 2]. FSW is categorized under the solid-state welding process and widely used to join non-ferrous materials such as aluminum, magnesium, and copper structures across many industries where a high strength to lightweight welds are required [3-7]. Compared to the conventional fusion welding process, it consumes lower energy, and no consumable materials such as electrodes and shielding gases are used. The application area of FSW is implemented in different industries, including aerospace, transportation, railway, shipbuilding, and other manufacturing industries due to its benefits such as a higher strength to weight ratio, corrosion resistance, and thermo-mechanical properties [8, 9]. Nowadays, the need for lightweight materials and their joining process is required in many manufacturing industries. Among these, magnesium alloy is a lightweight material, and its demand is rapidly increasing in the automotive and aerospace industry due to its low density and high specific strength. The material is approximately 30% lighter than aluminum and four times lighter than steel, with a density of 1.8 g/cm3 [10]. In addition to this, FSW is a suitable joining process for magnesium alloy materials because the weld takes place below the melting point of the base metal. Both materials and joining processes are compatible for imparting a sound weld; however, they need appropriate process parameters. [11] has studied the effect of process parameters of aluminum-magnesium alloy (6061-T6) material using the FSW process. Parameters are controlled by using the Taguchi-based GRA method. The result revealed that a combination of higher rotational and lower traverse speed imparts a higher hardness and tensile strength to the weld joint. Moreover, the authors confirmed that FSW is a suitable welding process for getting a higher mechanical property in 6061-T6 material. [12] has investigated the impact of welding speed on the microstructure and mechanical properties of dissimilar materials of AA2024 and AA6061 aluminum material using the FSW process. They obtained defect-free joints at a tool rotational speed of 710 rpm, welding speed of 28mm/min, and D/d ratio of 3. Moreover, according to the statistical analysis, the tool pin is the most influential parameter for attaining an excellent mechanical property. [13] are studied the mechanical properties of 6061-T6 AA using the FSW process. The parameters are optimized by Taguchi L9 orthogonal array method. The result showed that the highest tensile strength was recorded at a higher tool rotational speed of 1400 rpm, tilt angle 0o, and 100 mm/min welding speed. Recently, an Artificial Intelligence-based approach known as Machine Learning start to implement in various manufacturing sectors, including the Friction Stir Welding Process. A supervised Machine Learning classification-based approach was used to predict the mechanical properties of Friction Stir Welded Copper joints [14]. [15] determined the fracture location of dissimilar Friction Stir Welded joints using a supervised machine learning classification-based approach.

This work aims to find optimum process parameters of FSW for enhancing the tensile strength of the target material using the Machine Learning approach. The results showed that the Support Vector Machine (SVM) algorithm resulted in the highest accuracy score of 0.889. Furthermore, machine Vision-based algorithms have also been used in the Friction Stir Welding Process to determine surface defects and microstructure geometrical features analysis in Friction Stir Welded joints [16, 17].

## 2. Experimental Procedure

### 2.1 Materials and Setup

The material used in this investigation is a 5 mm 6061 Al alloy sheet with a butt joint configuration, and the chemical composition and mechanical properties of the material is summarized in Table 1 and 2 respectively. The specimen was prepared from a rolled sheet in a rectangular shape of 101.6 x 20 x 5 mm sliced by a hacksaw to diminish the residual stresses through the cutting processes. Cylindrical threaded, taper threaded, and tri-flute threaded tool pins see Fig. 1a, with a Ø 4.7 mm and Ø 15 mm pin and shoulder diameter, respectively, is used for welding the specimen. The cylindrical tool is made from H13 tool steels, the tri-flute tool is made from C40 steel, and the taper tool is also made from HSS tool steel materials. The tools' mechanical properties and chemical composition are depicted in Table 3-5. Throughout the experimentations, the plunge depth for providing proper axial load was 0.3 mm. Samples are fabricated illustrated in Fig. 1b using XHS7145 vertical CNC milling machining center. Tensile test samples based on ASTM-E8-04 standard demonstrated in Fig 1c. were retrieved from the fabricated joints utilizing VMBS 1610 band saw machine to evaluate the tensile strength of the weld joints. Then the tensile tests were carried out using the Bairoe computer-controlled electro-hydraulic universal testing machine. Tensile strength was measured triple times and take an average result. Fig. 1d shows the dog shape samples before and after the tensile test.

Table 1. AA6061 Chemical Composition

| Material | Mg | Si | Fe | Cr | Cu | AL |
|---|---|---|---|---|---|---|
| AA 6061 | 0.92% | 0.6% | 0.33% | 0.18% | 0.25% | 97.72% |

Table 2. AA6061 Mechanical Properties [18]

| Material | Yield strength, (MPa) | Tensile strength, (MPa) | Hardness, (HR) |
|---|---|---|---|
| AA 6061 | 276 | 310 | 40 |

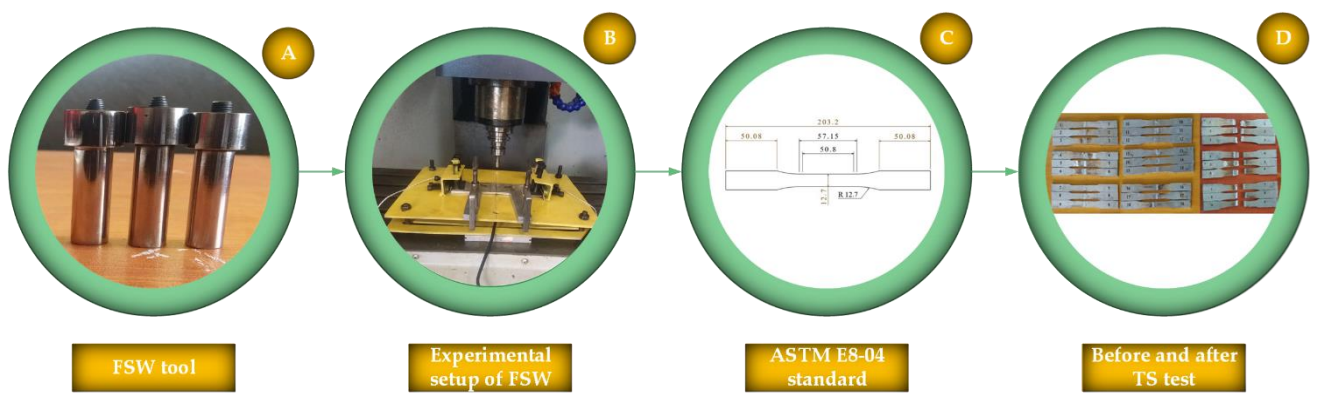

Fig. 1. Overall experimental setup of FSW process

## 2.2 Implementation of Machine Learning Algorithms

The process of implementing the machine learning algorithms is shown in Fig. 2. Firstly, the necessary Python libraries such as pandas, NumPy, seaborn, matplotlib, and seaborn are imported to the working environment. In the second step, the dataset was imported to the Jupyter notebook environment and further checked for missing values. Thirdly, exploratory data analysis is carried out to analyze the dataset to extract the main characteristics features using a graphical statistical approach as shown in Fig. 3.

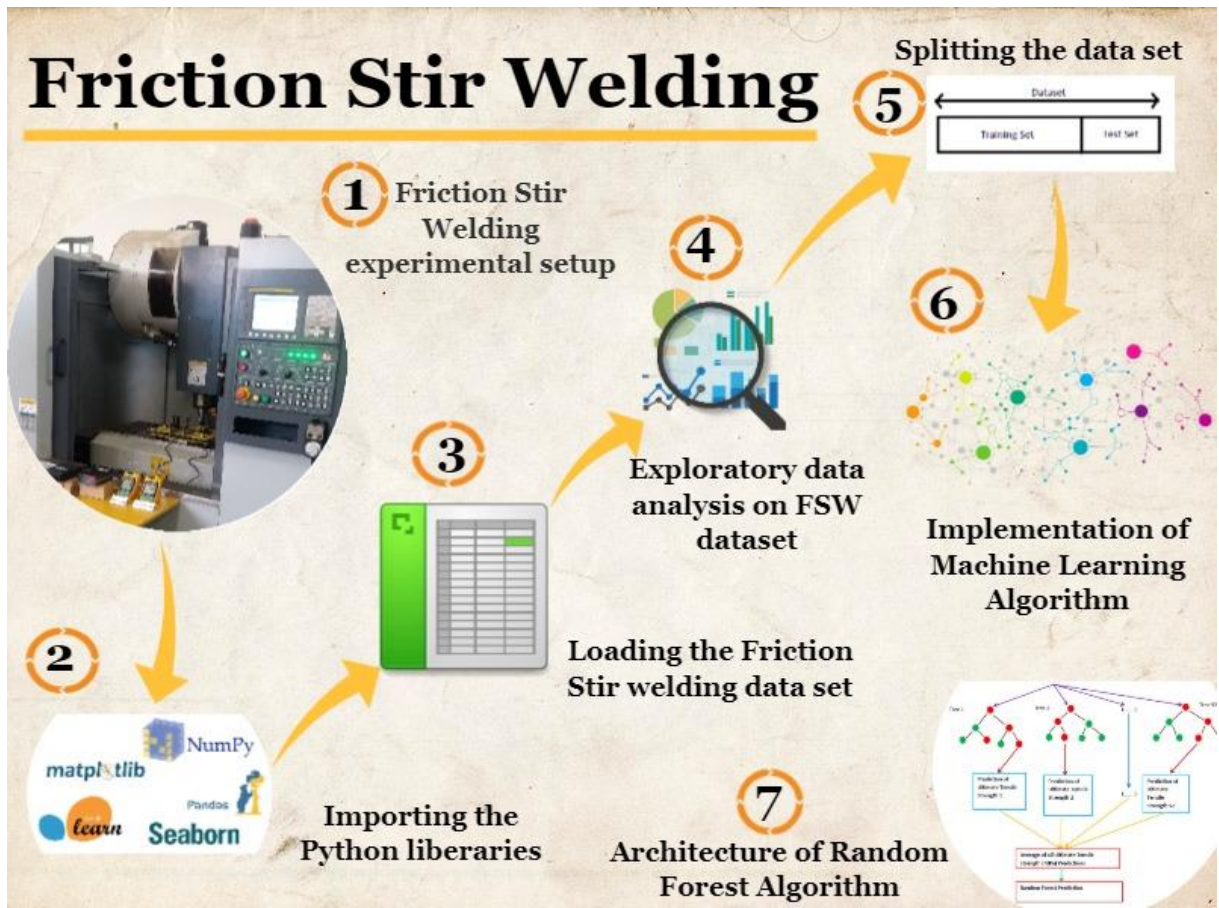

Fig.2. Schematic Representation of the implementation of Machine Learning algorithms

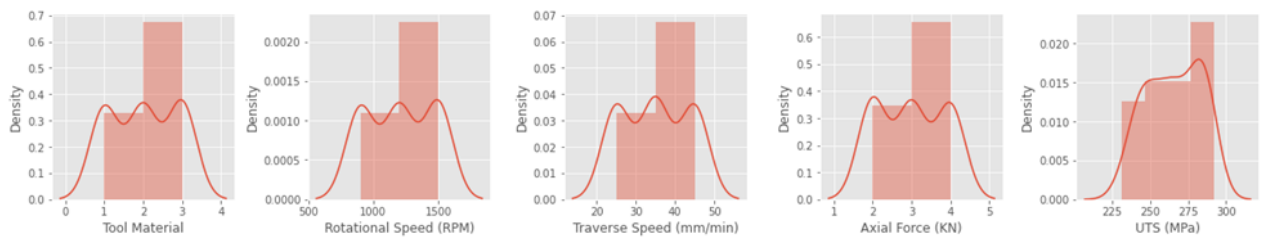

Fig.3. Exploratory data analysis on the given dataset

Fourthly, feature importance for each input feature is calculated as shown in Fig. 4. It is observed that the rotational tool speed (RPM) has the highest influence on the nature of UTS (MPa), followed by tool traverse speed (mm/min) and axial force (KN). It is also observed that the tool material does not influence the nature of UTS (MPa).

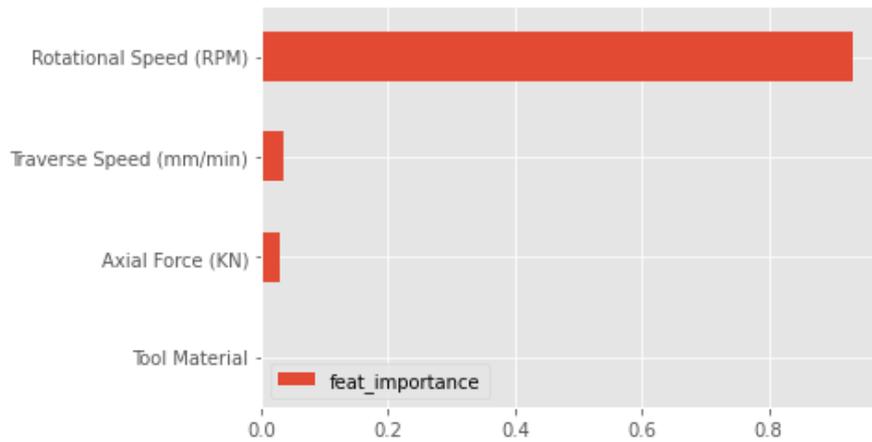

Fig. 4. Plot of feature importance

In the fifth step, splitting the dataset is done in an 80-20 ratio, i.e., 80 percent of the data is used for training purposes, and 20 percent of the data is used for testing purposes. In the last step, the value of the metrics features such as Mean Square Error, Mean Absolute Error, and coefficient of determination ($R^2$) are calculated for measuring the performance of the individual Supervised Machine Learning algorithms.

Table 3. Process parameters and their levels

| Exp. Nº | Tool material [Type] | Rotational speed [RPM] | Welding speed [mm/min] | Axial force [KN] | UTS [MPa] |
|---|---|---|---|---|---|
| 1 | H13 | 900 | 25 | 2 | 251 |
| 2 | H13 | 900 | 25 | 2 | 254 |
| 3 | H13 | 900 | 25 | 2 | 257 |
| 4 | H13 | 1200 | 35 | 3 | 264 |
| 5 | H13 | 1200 | 35 | 3 | 260 |
| 6 | H13 | 1200 | 35 | 3 | 268 |
| 7 | H13 | 1500 | 45 | 4 | 284 |
| 8 | H13 | 1500 | 45 | 4 | 284 |
| 9 | H13 | 1500 | 45 | 4 | 281 |
| 10 | H13 | 900 | 35 | 4 | 242 |
| 11 | H13 | 900 | 35 | 4 | 244 |
| 12 | H13 | 900 | 35 | 4 | 241 |
| 13 | H13 | 1200 | 45 | 2 | 264 |
| 14 | H13 | 1200 | 45 | 2 | 264 |
| 15 | H13 | 1200 | 45 | 2 | 260 |
| 16 | H13 | 1500 | 25 | 3 | 288 |
| 17 | H13 | 1500 | 25 | 3 | 288 |
| 18 | C40 | 1500 | 25 | 3 | 286 |
| 19 | C40 | 900 | 45 | 3 | 238 |
| 20 | C40 | 900 | 45 | 3 | 231 |
| 21 | C40 | 900 | 45 | 3 | 236 |
| 22 | C40 | 1200 | 25 | 4 | 271 |
| 23 | C40 | 1200 | 25 | 4 | 268 |
| 24 | C40 | 1200 | 25 | 4 | 273 |
| 25 | C40 | 1500 | 35 | 2 | 281 |
| 26 | C40 | 1500 | 35 | 2 | 278 |
| 27 | C40 | 1500 | 35 | 2 | 280 |
| 28 | C40 | 900 | 25 | 2 | 248 |
| 29 | C40 | 900 | 25 | 2 | 248 |
| 30 | C40 | 900 | 25 | 2 | 245 |
| 31 | C40 | 1200 | 35 | 3 | 258 |
| 32 | C40 | 1200 | 35 | 3 | 257 |
| 33 | C40 | 1200 | 35 | 3 | 254 |
| 34 | C40 | 1500 | 45 | 4 | 281 |
| 35 | HSS | 1500 | 45 | 4 | 286 |
| 36 | HSS | 1500 | 45 | 4 | 285 |
| 37 | HSS | 900 | 35 | 4 | 248 |
| 38 | HSS | 900 | 35 | 4 | 246 |
| 39 | HSS | 900 | 35 | 4 | 247 |
| 40 | HSS | 1200 | 45 | 2 | 266 |
| 41 | HSS | 1200 | 45 | 2 | 264 |
| 42 | HSS | 1200 | 45 | 2 | 269 |
| 43 | HSS | 1500 | 25 | 3 | 291 |
| 44 | HSS | 1500 | 25 | 3 | 292 |
| 45 | HSS | 1500 | 25 | 3 | 291 |
| 46 | HSS | 900 | 45 | 3 | 239 |
| 47 | HSS | 900 | 45 | 3 | 242 |
| 48 | HSS | 1200 | 25 | 4 | 276 |
| 49 | HSS | 1200 | 25 | 4 | 274 |
| 50 | HSS | 1500 | 35 | 2 | 286 |
| 51 | HSS | 1500 | 35 | 2 | 285 |
| 52 | HSS | 1500 | 35 | 2 | 285 |

## 3. Results and Discussion

### 3.1 Decision Tree Algorithm

Decision Tree is a greed-based non-parametric machine learning algorithm used to predict the target variable, i.e., UTS (MPa), using some decision rules in the present study. The Decision Tree algorithm builds the model in the tree architecture by splitting the dataset into various subsets, resulting in its final form in decision nodes and leaf nodes, as seen in Fig. 5.

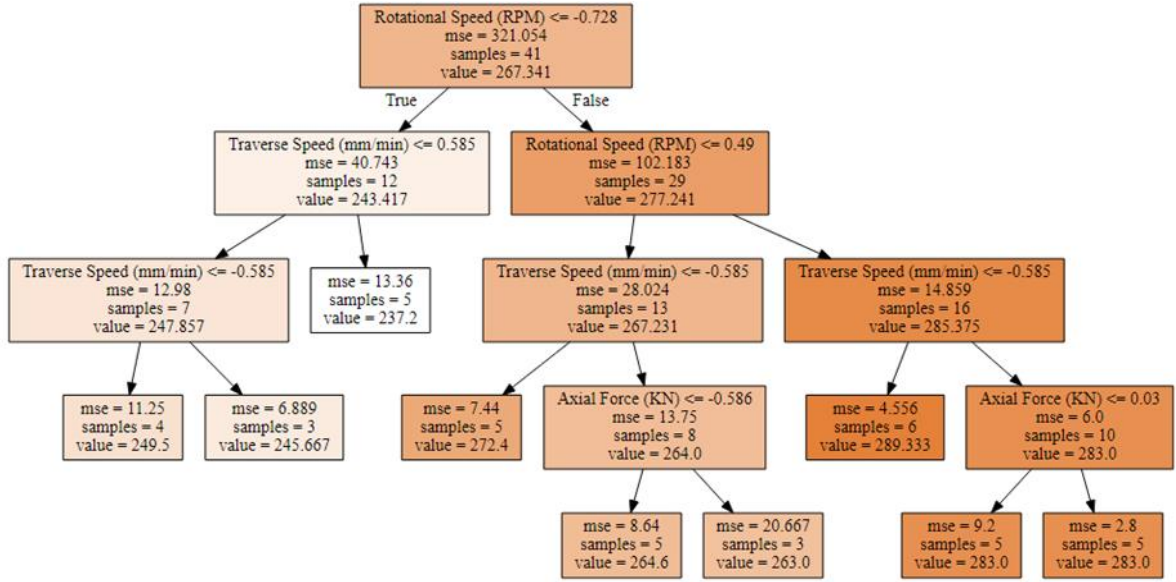

Fig. 5. Decision Tree architecture in the present work

Decision Tree architecture is formed by the recursive partitioning methodology, which starts from the first parent, i.e., the root node, which split into right and left child nodes, and they further split themselves as a parent node resulting in other child nodes. Based on the largest Information Gain (IG), the dataset is split into features starting from the root node, and thus this process is repeated iteratively. The main objective is to maximize the Information Gain (IG), as shown in Equation 1, splitting the nodes at most informative features.

$$IG(D_p, f) = I(D_p) - \left(\frac{N_{Left}}{N_p} I(D_{Left}) + \frac{N_{Right}}{N_p} I(D_{Right})\right) \qquad (1)$$

Where $D_p, D_{Left}, D_{Right}$ are the dataset of the child and parent nodes, $N_{Right}, N_{Left}$ are number of samples in child nodes, $N_p$ is the total number of samples at the parent node and $I$ is the impurity measure. Table 4 shows the performance of the Decision Tree algorithm evaluated by the measurement of the metric features.

Table 4. Metric features evaluation of Decision Tree Algorithm

| Mean Square Error | Mean Absolute Error | $R^2$ Value |
|---|---|---|
| 19.684 | 3.569 | 0.894 |

## 3.2 Gradient Boosting Algorithm

Gradient Boosting algorithm is an ensemble model that combines the weal learners or weak predictive models to predict the continuous value further. This approach can be used for both classification and regression purposes. In this work, it is used as a regression algorithm for predicting the UTS (MPa). The multiple decision trees of a constant size as weak predictive models are summed up to build an additive model. To predict the negative gradients of the samples in the dataset, the decision tree-based estimators are fitted. In gradient boosting algorithm, M stages are considered, and some imperfect model $F_m$ is present at each stage of $m (1 \leq m \leq M)$ of gradient boosting. So, a new estimator $h_m(x)$ is added to the algorithm for improving the value of $F_m$ it is shown in Equation 2.

$$F_{m+1}(x) = F_m(x) + h_m(x) \qquad (2)$$

Table 5 shows the performance evaluation of the Gradient Boosting Algorithm.

Table 5. Metric features evaluation of Gradient Boosting Algorithm

| Mean Square Error | Mean Absolute Error | $R^2$ Value |
|---|---|---|
| 19.939 | 3.591 | 0.893 |

It is observed that the MSE and MAE obtained from Gradient Boosting Algorithm is higher than those of the Decision Tree Algorithm while $R^2$ value is slightly lower than the Decision Tree Algorithm.

### 3.3 Random Forest Algorithm

Random Forest algorithm is a supervised machine learning algorithm constituted from decision trees and can be used to solve regression and classification-based problems. The solution to the complex problem is met by ensemble learning in which various classifiers are combined. The predicted output is generated by the outcome resulted from the decision trees by taking the mean or average of the output yielded from the individual decision trees, as shown in Figure 6. Table 6 shows the performance metrics evaluation of the Random Forest Algorithm.

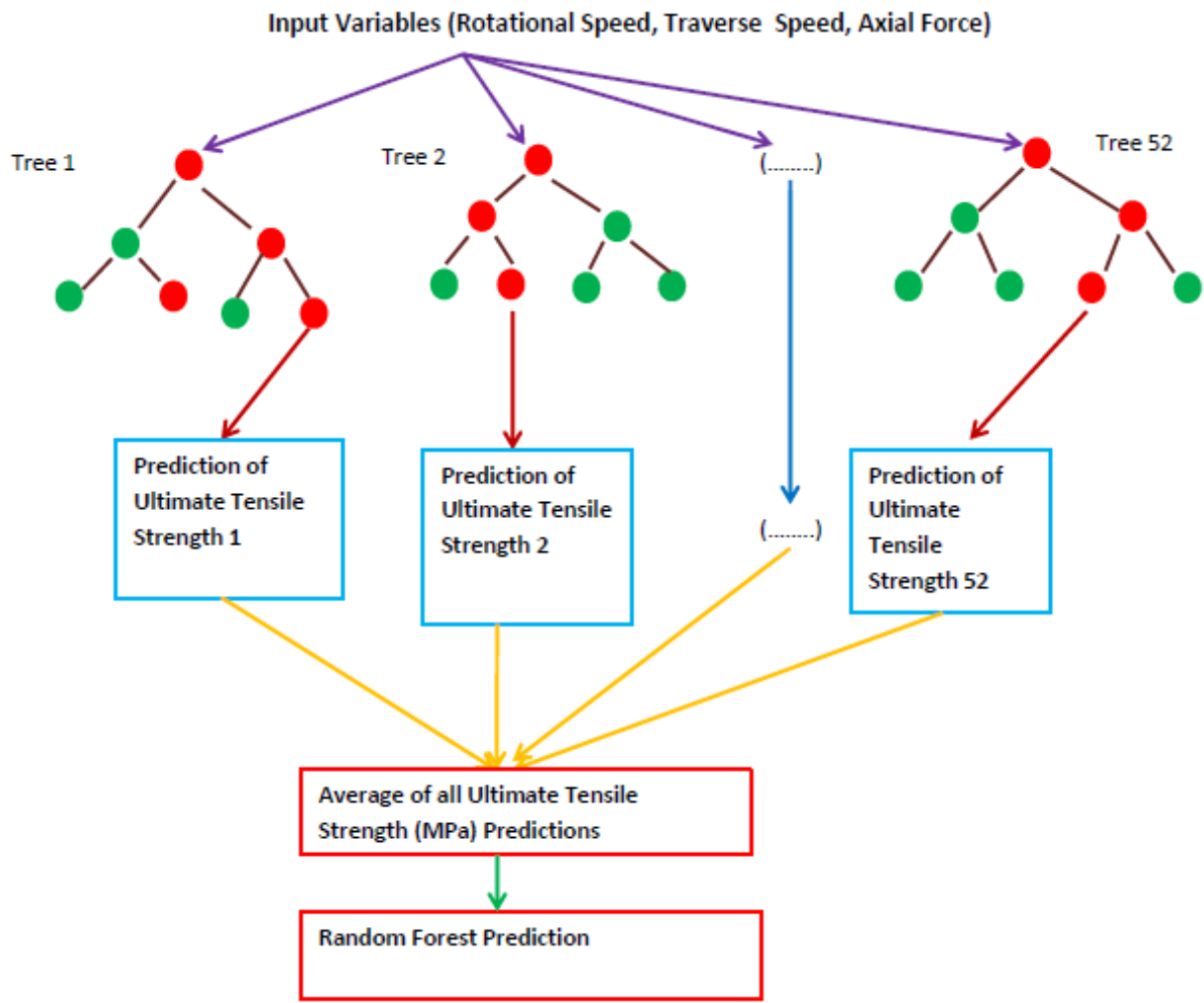

Fig. 6. Architecture of Random Forest algorithm in present work

Table 6. Metric features evaluation of Random Forest Algorithm

| Mean Square Error | Mean Absolute Error | $R^2$ Value |
|---|---|---|
| 19.079 | 3.717 | 0.926 |

It is observed that the Random forest results the best fit by yielding the highest $R^2$ value of 0.926 in comparison to the other two algorithms. It can be concluded that the Random Forest algorithm yields the highest level of accuracy in comparison to the decision trees for predicting the output, i.e., UTS (MPa). It is observed that each implemented ML classification model has 100 % accuracy for the classification of the welding efficiency of friction stir welded joints.

## 4. Conclusion

This paper conducted the test at ambient temperature for similar materials of Friction Stir welded. The maximum tensile strength was 292 MPa obtained from a tapered pin profile of the HSS tool at a rotational speed of 1500 rpm with a welding speed of 25 mm/min and axial force of 3KN. Correspondingly, the minimum tensile strength value of 231 MPa was observed from a tri-flute threaded profile of the C40 tool at a rotational speed of 900 rpm with a traverse speed of 45 mm/min axial force of 3KN. The joint efficiency was reached 94.2 % of the base metal. The result shows that the tensile strength increases, the traverse speed of the tool decreases, and the rotational speed of the tool increases because the lower welding speed and higher tool rotational speed can impart sufficient heat for welding the parent metal.

In the present work, an experimental study has been optimized through supervised machine learning regression-based algorithms such as Decision Trees, Gradient Boosting Algorithm, and Random Forest Algorithm are developed and executed using Python programming to finding the effect of the high mechanical properties of AA6061 AA joints fabricated through the friction stir welding process. The obtained results can be summarized as follows:

- The highest tensile strength (UTS) of 292 MPa was obtained at a parameter setting of the rotational speed of 1500 rpm, welding speed of 25 mm/min, the axial force of 3 KN, and a taper threaded tool pin.

- In the dataset, firstly, there were four input features as Tool Material, Tool Rotational Speed (RPM), Tool Traverse Speed (mm/min), and Axial force (KN). However, the feature importance results showed that the Tool Material does not affect the output feature, i.e., UTS of the FS welded joints, which led to the dropping of the tool material input feature.

- The obtained results showed that the Random Forest algorithm resulted in the highest coefficient of determination of 0.926 compared to the other two algorithms, which means the predicted results are very close to the experimental results.

The future scope of this work is to implement Quantum Machine Learning based algorithms and further compare the results obtained from the conventional Machine Learning Algorithms.

## Author Contributions

**Competing Interests**

The authors declare that they have no competing personal and financial interests.

**Acknowledgment**

The authors would like to acknowledge the School of Research and Graduate Studies, Bahir Dar Institute of Technology, for their support.